\documentclass[letterpaper, 10 pt, conference]{ieeeconf}  

\IEEEoverridecommandlockouts                              

\usepackage{amsmath,amsfonts}
\usepackage{algorithmic}
\usepackage{algorithm}
\usepackage{array}
\usepackage{textcomp}
\usepackage{stfloats}
\usepackage{url}
\usepackage{verbatim}
\usepackage{graphicx}
\usepackage{hyperref}
\usepackage{cite}
\usepackage{multirow}%
\usepackage{xcolor}%
\usepackage{booktabs}%
\usepackage{colortbl}
\usepackage[table]{xcolor}
\usepackage{subcaption}
\usepackage{pifont}
\usepackage{circledsteps}
\newcommand{\supcirc}[1]{\textsuperscript{\Circled{#1}}}
\usepackage{svg}

\newcommand{\cmark}{\textcolor{green!55!black}{\ding{51}}}
\newcommand{\xmark}{\textcolor{red!70!black}{\ding{55}}}

\definecolor{mAPgreenDark}{HTML}{C8E6C9}      %
\definecolor{mAPgreenLight}{HTML}{E8F5E9}     %
\definecolor{stableBlueDark}{HTML}{BBDEFB}    %
\definecolor{stableBlueLight}{HTML}{E3F2FD}   %

\title{\LARGE \bf
Pro-Bench: Prompt-Robust Open-Vocabulary Visual Grounding Across Real-World Heterogeneous Environments%
}

 \author{Linus Nwankwo$^{*1}$; Muslim Alaran$^{1}$; Christian Rauch$^{1}$; Stanley Chukwuebuka Obilikpa$^{2}$; Elmar Rueckert$^{1}$
 \thanks{This work is supported by the ``MINEVIEW'' Project (\#FO999927835), funded by the Rep. of Austria, Fed. Min. of Env., Innovation and Tech.}
 \thanks{$^{1}$Chair of Cyber-Physical Systems, Montanuniversität Leoben, Austria.
        }%
\thanks{$^{2}$Department of Design Engineering and Mathematics, Middlesex University, London, UK.
         {*Corresponding author:~\tt\small linus.nwankwo@unileoben.ac.at}
         }%
 }

\begin{document}

\maketitle
\thispagestyle{empty}
\pagestyle{empty}

\begin{abstract}
    Open-vocabulary visual grounding enables robots to localise task-relevant entities from natural-language queries without dependence on predefined perceptual taxonomies. However, existing benchmarks largely rely on short category labels and web-scraped imagery, leaving it unclear whether open-vocabulary models can robustly ground diverse queries and visual conditions under real deployments. We introduce \textbf{Pro-Bench}, a prompt-conditioned benchmark for open-vocabulary visual grounding in heterogeneous, real-world environments. Pro-Bench includes $13k+$ RGB frames from independent robotic domains (subterranean, industrial, indoor, outdoor, urban), with $74.5k$ manual instance annotations and $515$ target queries covering categorical, attributive, relational, affordance, state, part-whole, negative, and compositional semantics. We benchmarked $16$ open-vocabulary model configurations in strict zero-shot inference, measuring localisation accuracy across IoU thresholds, end-to-end inference latency, prompt-induced performance variation, and target recovery consistency. 
    Our results show that prompt-robustness is strongly architecture-dependent. Most model configurations ($10/16$) perform best with short category labels, whereas free-form queries yield the highest accuracy for only one. Moreover, similar aggregate mAP can conceal substantial differences in consistent target recovery across reformulations. Pro-Bench enables systematic evaluation of these gaps and supports prompt-robust visual grounding. Pro-Bench:~\url{https://pro-bench.github.io/}.
\end{abstract}

\section{Introduction}\label{sec:intro}
    Recently, there has been a growing interest in deploying open-vocabulary models (OVMs)~\cite{minderer2022simple,minderer2023scaling, liu2024grounding, zhou2022detecting, carion2025sam,cheng2024yolo, wang2025yoloe} across autonomous robot platforms.
\begin{figure*}[htp]
    \centering
    \includegraphics[width=1.0\linewidth]{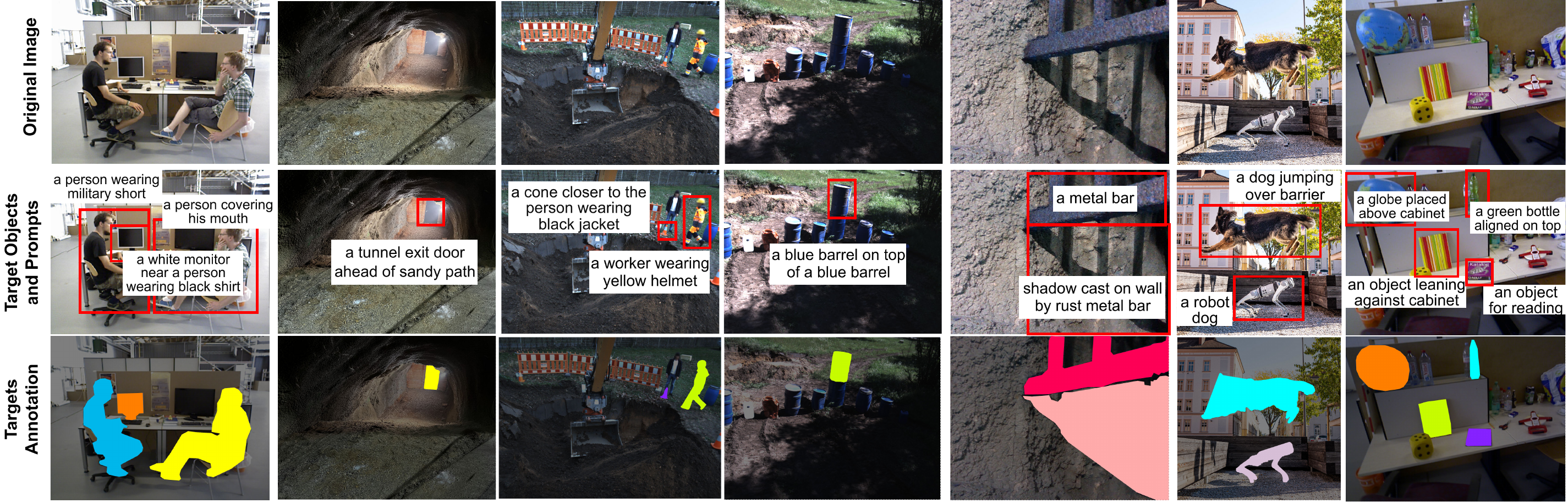}
    \caption{Representative Pro-Bench sample.  Images span diverse robotic domains (Section~\ref{sec:dataset}) with varied illumination, viewpoints, and object categories. The prompt design and category definitions are described in Section~\ref{sec:prompt_design}. We provide our ground-truth annotations in multiple formats, e.g., COCO~\cite{lin2014microsoft}, suitable for visual grounding applications.}
    \label{fig:dataset_overview}
\vspace*{-1.15\baselineskip}
\end{figure*}
    Unlike closed-vocabulary models~\cite{varghese2024yolov8,tian2026yolov12,ren2015faster,liu2016ssd,carion2020end}, OVMs, in principle, enable visual recognition conditioned on natural-language prompts, allowing target objects to be specified at inference time rather than being restricted to fixed categories defined during the model training~\cite{wu2023cora,li2025ovtr,bangalath2022bridging, xie2023described}. This capability is particularly important for robots that must operate in task-driven environments such as subterranean exploration~\cite{ohradzansky2022multi,kulkarni2022autonomous,biggie2023flexible}, industrial inspection~\cite{gu2024anomalygpt,ge2020towards}, search-and-rescue operations~\cite{panagopoulos2024selective,shree2021exploiting}, and language-conditioned human-robot collaboration~\cite{nwankwo2025beyond, ahn2022can}, where the target space (i.e., task-relevant objects) is inherently open-ended, context- and mission-dependent, and cannot be fully enumerated a priori. Instead, the operator's natural language prompt dynamically determines the perception objective.
    
    However, standard benchmarks~\cite{lin2014microsoft,mao2023coco,gupta2019lvis,wang2023v3det,shao2019objects365,kuznetsova2020open,li2022elevater} largely evaluate OVMs' visual reasoning against short, atomic category labels (e.g., \emph{person}, \emph{cup}, \emph{cat}, etc.), which often do not fully encode attributes, spatial relations, compositions, implicit affordances, physical state, or constituent parts, typically required in practice, and on predominantly web-sourced or synthetic image modalities. This raises concerns about the current OVMs' generalisability beyond such categorical labels, and the robustness to diverse environmental conditions (e.g., low illumination, occlusion, unusual viewpoints, sensor artefacts, weather, and domain-specific objects), which are characteristic of real-time embodied robotic operation.
    
    Furthermore, prompt robustness, which quantifies the extent to which linguistic variations in a prompt affect model predictions, remains another under-explored challenge in the current OVM benchmarks. Natural language admits multiple expressions for the same physical target. For example, \emph{``a tunnel ventilation fan on the ceiling''}, \emph{`` a photo of \{tunnel ventilation fan\}''}, or \emph{``ventilation fan''} may all denote the same instance but may yield inconsistent target detections. We contend that prompt robustness is crucial for safe human-robot interaction, to enable robots to infer human intent despite variations, ambiguity, or noise in the task instructions.
    
    Motivated by these challenges, we propose Pro-Bench, a prompt-conditioned benchmark for open-vocabulary visual grounding under conditions representative of real-world robotic deployment. Pro-Bench combines robot-relevant image modalities with instance-level annotations and semantically diverse target-grounded prompts, and augments each target with categorical and template-wrapped reformulations. Fig.~\ref{fig:dataset_overview} shows representative scenes, prompt types, and target annotations. Accordingly, our contributions are as follows:
\begin{itemize}
    \item \textbf{Prompt-rich benchmark.} We introduce Pro-Bench, a benchmark for prompt-conditioned open-vocabulary visual grounding that encompasses $13k+$ RGB frames captured across heterogeneous environments, $74.5k$ fine-grained human-annotated target instances, and $515$ prompt-defined queries, with $8$ diverse semantic prompt types, including compositional multi-constraint queries.
    \item \textbf{Prompt robustness protocol.} We propose a prompt-robust protocol that quantifies the impact of query reformulation on OVMs' predictions. Beyond aggregate performance variation, we introduce prompt-consistent recall and localisation consistency, which measure whether the same ground-truth target is recovered across reformulations and whether its localisation remains spatially consistent across prompt-conditioned predictions.
    \item \textbf{Zero-shot benchmark of representative OVMs.} We benchmarked $16$ OVM configurations on the Pro-Bench dataset in a strict zero-shot setting, without Pro-Bench-specific fine-tuning, and characterised their real-time Pareto frontier by evaluating detection accuracy, prompt-robustness, and accuracy–latency trade-offs across multiple model scales.
\end{itemize}
\textbf{Open-source contribution.} The Pro-Bench dataset, annotations, utilities, and per-model configurations will be released to support reproducible benchmarks and future extensions.


\section{Related Work}\label{sec:related}
Numerous benchmarks have been proposed over the years to evaluate OVMs, each with a specific focus and limitations. We review them along three attributes: vocabulary scale, visual-domain diversity, and linguistic richness, and group them into large-scale categorical, domain-diverse, and fine-grained/referential benchmarks. Table~\ref{tab:benchmark_comparison} summarises their key characteristics against Pro-Bench. We restrict the comparison to benchmarks that evaluate text-prompted box or mask localisation, the task performed by OVMs.

{\scriptsize
\begin{table*}[htp]
\centering
\caption{Pro-Bench versus representative open-vocabulary detection benchmarks. \textbf{Cat}egorical, \textbf{Attr}ibutive, \textbf{Rel}ational, \textbf{Comp}ositional, \textbf{Aff}ordance, \textbf{Sta}te, \textbf{Part}-whole, and \textbf{Neg}ative are the prompt annotation taxonomies, see Fig.~\ref{fig:envobench-pipeline}, panel\supcirc{3}. \textbf{Sens}itivity indicates whether robustness to alternative prompt formulations is evaluated. \cmark~denotes that the corresponding capability is explicitly evaluated in the benchmark's language-conditioned target/query label space, and \xmark~indicates otherwise.}
\label{tab:benchmark_comparison}
\small
\setlength{\tabcolsep}{1.8pt}
\renewcommand{\arraystretch}{1.05}
\begin{tabular}{@{}l l c c c c c c c c c c c c c c@{}}
\toprule
& \textbf{Benchmark} & \textbf{Scale} & \textbf{AnIns} & \textbf{Lang/Voc} & \textbf{Domain} & \textbf{AnTyp}
& \textbf{Cat} & \textbf{Attr} & \textbf{Rel} & \textbf{Comp} & \textbf{Aff} & \textbf{Sta} & \textbf{Part} & \textbf{Neg} & \textbf{Sens} \\
\cmidrule{2-16}
\multirow{5}{*}{\rotatebox[origin=c]{90}{\scriptsize\bfseries\shortstack{Large-scale cat.\\ detection}}}
& COCO~\cite{lin2014microsoft} & 118k & 860k & 80 cls. & Web & B+M
& \cmark & \xmark & \xmark & \xmark & \xmark & \xmark & \xmark & \xmark & \xmark \\
& LVIS~\cite{gupta2019lvis} & 100k & 1.27M & 1{.}2k cls. & Web & B+M
& \cmark & \xmark & \xmark & \xmark & \xmark & \xmark & \xmark & \xmark & \xmark \\
& Objects365~\cite{shao2019objects365} & 638k & 10.1M & 365 cls. & Web & B
& \cmark & \xmark & \xmark & \xmark & \xmark & \xmark & \xmark & \xmark & \xmark \\
& V3Det~\cite{wang2023v3det} & 243k & 1.8M & 13{,}204 cls. & Web & B
& \cmark & \xmark & \xmark & \xmark & \xmark & \xmark & \xmark & \xmark & \xmark \\
& Open Img.~\cite{kuznetsova2020open} & 1.9M & 15.4M & 600 cls. & Web & B
& \cmark & \xmark & \xmark & \xmark & \xmark & \xmark & \xmark & \xmark & \xmark \\
\cmidrule{2-16}
\multirow{4}{*}{\rotatebox[origin=c]{90}{\scriptsize\bfseries\shortstack{Domain\\diverse}}}
& ODinW~\cite{li2022elevater} & 132k & 1.07M & 314 concepts & Multi & B
& \cmark & \xmark & \xmark & \xmark & \xmark & \xmark & \xmark & \xmark & \xmark \\
& R.Flow 100~\cite{ciaglia2022roboflow} & 224k & 1.32M & 805 cls. & Multi & B
& \cmark & \xmark & \xmark & \xmark & \xmark & \xmark & \xmark & \xmark & \xmark \\
& LAE-80C~\cite{pan2025locate} & 3.6k & 86.6k & 80 cls. & Aerial/Sat. & B
& \cmark & \xmark & \xmark & \xmark & \xmark & \xmark & \xmark & \xmark & \xmark \\
& RefDrone~\cite{sun2026refdrone} & 8.5k & 63.7k & 17.9k expr. & Aerial & B
& \cmark & \cmark & \cmark & \cmark & \xmark & \xmark & \xmark & \xmark & \xmark \\
\cmidrule{2-16}
\multirow{9}{*}{\rotatebox[origin=c]{90}{\scriptsize\bfseries\shortstack{Fine-grained and\\referential}}}
& RefCOCO~\cite{yu2016modeling} & 20k & 50k & 142k expr. & Web & B+M
& \cmark & \cmark & \cmark & \cmark & \xmark & \xmark & \xmark & \xmark & \xmark \\
& gRefCOCO~\cite{liu2023gres} & 20k & 60.3k & 278k expr. & Web & B+M
& \cmark & \cmark & \cmark & \cmark & \xmark & \xmark & \xmark & \cmark & \xmark \\
& OVAD~\cite{bravo2023open} & 2k & 14.3k & 80 cls./117 attr. & Web & B
& \cmark & \cmark & \xmark & \xmark & \xmark & \xmark & \xmark & \xmark & \xmark \\
& FG-OVD~\cite{bianchi2024devil} & suite & varies & varies & Web & B
& \cmark & \cmark & \xmark & \cmark & \xmark & \xmark & \cmark & \xmark & \xmark \\
& 3F-OVD~\cite{liu2025fine} & 145.8k & 676.5k & 105 cls./719 sub. & Multi & B+C
& \cmark & \cmark & \xmark & \cmark & \xmark & \xmark & \xmark & \xmark & \xmark \\
& OmniLabel~\cite{schulter2023omnilabel} & 25k & 340k & $>$ 30k desc. & Web & B
& \cmark & \cmark & \cmark & \cmark & \xmark & \xmark & \xmark & \cmark & \xmark \\
& D$^{3}$~\cite{xie2023described} & 10.6k & 18.5k & 422 desc. & Web & B+M
& \cmark & \cmark & \cmark & \cmark & \xmark & \xmark & \xmark & \cmark & \xmark \\
& OVDEval~\cite{yao2024evaluate} & suite & NF & varies & Web & B
& \cmark & \cmark & \cmark & \cmark & \xmark & \xmark & \xmark & \cmark & \xmark \\
& RoboRefIt~\cite{lu2023vl} & 10.9k & NF & 50.8k expr. & Robot & B+M
& \cmark & \cmark & \cmark & \cmark & \xmark & \xmark & \xmark & \xmark & \xmark \\
\cmidrule{2-16}
\rowcolor{green!6}
& \textbf{Pro-Bench (ours)} & 13k & 74.5k & 515 mixed & \textbf{Multi} & \textbf{B+M}
& \cmark & \cmark & \cmark & \cmark & \cmark & \cmark & \cmark & \cmark & \cmark \\
\bottomrule
\end{tabular}
\par\smallskip
{\footnotesize
\textcolor{blue}{Legend}:
Scale $\rightarrow$ Number of images;
Lang/Voc $\rightarrow$ Size and form of the linguistic supervision, e.g., class categories (cls.), attributes (attr.), referring expressions (expr.), descriptions (desc.), and mixed includes all;
AnIns$\rightarrow$ Instance annotations;
NF$\rightarrow$Not found in the corresponding paper;
Robot$\rightarrow$ Robot-mounted/manipulation-scene capture;
AnTyp$\rightarrow$ Annotation type:
B$\rightarrow$Bounding box,
M$\rightarrow$Mask (incl.\ polygon), and
C$\rightarrow$Caption.
}
\end{table*}
}

\textbf{Large-scale categorical detection.}
    COCO~\cite{lin2014microsoft}, LVIS~\cite{gupta2019lvis}, Objects365~\cite{shao2019objects365}, V3Det~\cite{wang2023v3det} and Open Images V4~\cite{kuznetsova2020open} provide broad category coverage for visual recognition. However, their annotations remain largely class-centric, where target objects are often labelled with canonical names or short noun phrases. Thus, they do not assess whether OVMs can resolve richer semantic constraints common in practice, such as attributes, spatial relations, physical state, etc.
    
\textbf{Domain-diverse.}
    ODinW~\cite{li2022elevater}, R.Flow~100~\cite{ciaglia2022roboflow}, and LAE-80C~\cite{pan2025locate} extend visual-domain coverage beyond everyday web imagery, but primarily retain category-level target supervision. RefDrone~\cite{sun2026refdrone} introduces referring expressions, though only for aerial viewpoints.
 
\textbf{Fine-grained and referential.}
    These benchmarks~\cite{yu2016modeling, bravo2023open,bianchi2024devil, liu2025fine} extend category-level supervision via referring expressions, visual attributes, and sub-category descriptions. Similarly, gRefCOCO~\cite{liu2023gres}, OmniLabel~\cite{schulter2023omnilabel}, D$^{3}$~\cite{xie2023described}, and OVDEval~\cite{yao2024evaluate} also include negative or no-target queries. Although these datasets provide multiple expressions per target, they do not evaluate whether predictions remain consistent under target-preserving reformulation, or span heterogeneous, multi-domain robot deployments.

\section{The Pro-Bench}\label{sec:benchmark} 
\begin{figure*}[ht]
        \centering
        \includegraphics[width=1.0\linewidth]{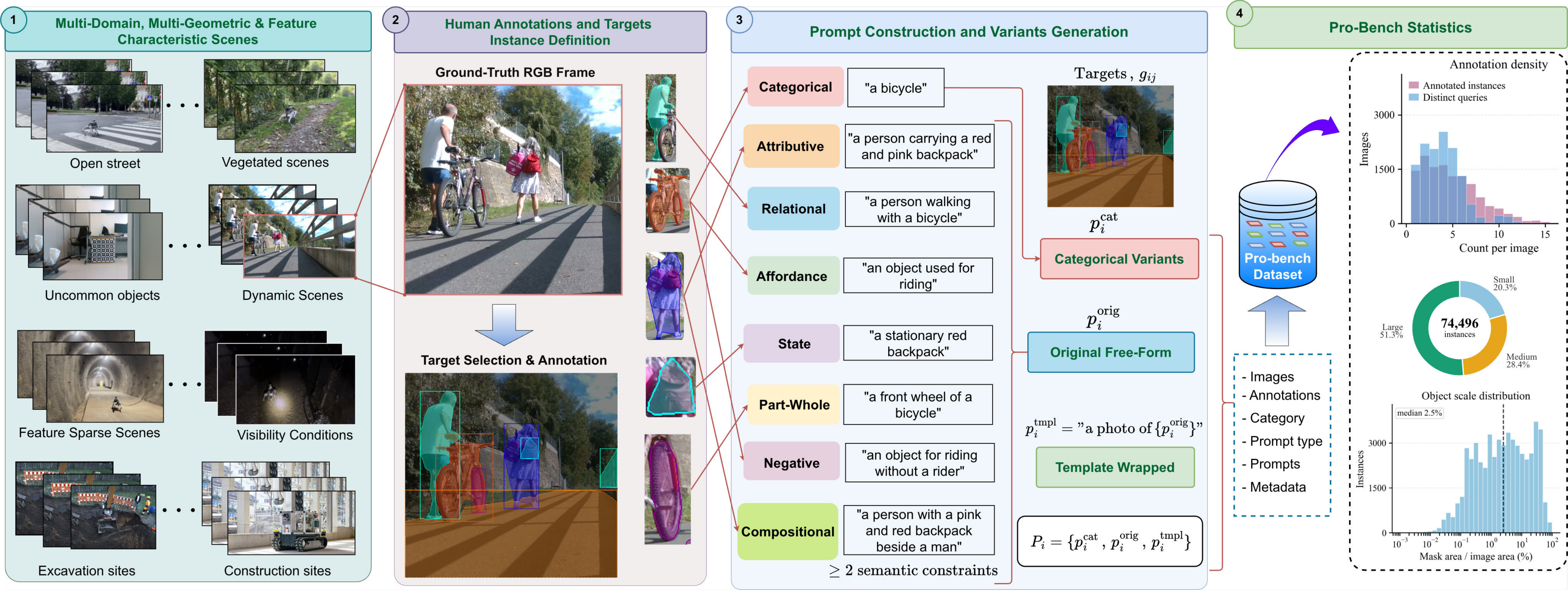}
        \caption{Overview of the Pro-Bench construction pipeline and the data statistics. Given RGB frames curated from multi-domain robotic datasets that depict realistic robot operating scenarios\supcirc{1}, we provide annotations for task-relevant targets with fine-grained polygon-based masks\supcirc{2}. Each target is associated with a human-defined free-form query and at least one semantic prompt category, from which we derive the categorical and template-wrapped variants for prompt-sensitivity evaluation\supcirc{3}. The final panel\supcirc{4} summarises the resulting Pro-Bench dataset statistics: $74.5k$ instance annotations distributed across different object sizes (small, medium, and large), over $13k$ images, and $515$ prompts.}
        \label{fig:envobench-pipeline}
    \vspace*{-1.15\baselineskip}
\end{figure*}

    We construct Pro-Bench from public and self-curated real-world robotic datasets spanning heterogeneous operating conditions (Section~\ref{sec:dataset}). Each selected RGB frame contains task-relevant instances annotated at fine spatial granularity and associated with target-grounded natural-language queries (Section~\ref{sec:prompt_design}). The queries span eight semantic prompt types (Fig.~\ref{fig:envobench-pipeline}, panel\supcirc{3}) designed to reflect how human operators may describe, identify, and refer to objects in real-time robotic missions. Fig.~\ref{fig:envobench-pipeline} summarises the Pro-Bench construction pipeline and dataset statistics.
    
\subsection{Pro-Bench Datasets and Domains}\label{sec:dataset}
    Table~\ref{tab:benchmark_comparison} (bottom) shows the Pro-Bench dataset scale and instance annotations. In addition to our self-collected data, we aggregate RGB frames from multiple robotic domains to capture environmental diversity and deployment-relevant distribution shifts. We consider only datasets recorded in situ with real robot sensing platforms (e.g., RGB-D and monocular cameras), and exclude non-robot, web-scraped, and synthetic image modalities to retain deployment-relevant variation in illumination, viewpoint, occlusion, scene structure, and object appearance. We group the corresponding datasets by their environment attributes as follows:
    
\textbf{Subterranean and industrial environments.}
    These include tunnels, quarry operations, and mining scenarios characterised by low illumination, constrained geometry, and domain-specific infrastructure (e.g., rails, ventilation systems).
    We sourced the corresponding RGB frames from the DARPA~\cite{rouvcek2019darpa} and EnvoDat~\cite{nwankwo2025envodat} subterranean challenge datasets, the Hilti-SLAM~\cite{nair2024hilti} underground and construction dataset, and the GOOSE-Ex~\cite{hagmanns2025excavating} excavation datasets.

\textbf{Outdoor and urban environments.}
    This includes forest trails, unstructured terrain, and urban street scenes with dynamic entities and infrastructure (e.g., vehicles, pedestrians). These scenes exhibit strong lighting/viewpoint changes (e.g., shadow formations caused by swaying foliage). We obtained these data from EnvoDat~\cite{nwankwo2025envodat} outdoor scenes, the GOOSE-Ex~\cite{hagmanns2025excavating} forest trails and field paths, the KITTI~\cite{geiger2013vision}, the MCD~\cite{nguyen2024mcd} multi-campus, and our self-collected campus data.

\textbf{Indoor and laboratory environments.}
    These include office, laboratory, and tabletop scenes with comparatively structured layouts but frequent visual clutter and multiple visually similar objects. We derive these data from the TUM RGB-D benchmark~\cite{sturm2012benchmark}, the EnvoDat~\cite{nwankwo2025envodat} indoor scenes, and our in-house collected scenes. 

\subsection{Prompt Taxonomy and Construction}\label{sec:prompt_design}
    A defining characteristic of Pro-Bench is its use of free-form natural-language prompts, such that target instances can be queried using semantically diverse natural-language expressions, rather than only canonical class names. For each annotated target (Section~\ref{sec:target-annotation}), we construct an original prompt $p^{\text{orig}}_{i}$ and assign at least one prompt type according to the semantic information required to identify its referent: \emph{categorical} (class identity), \emph{attributive} (intrinsic properties), \emph{relational} (relations to other scene entities),  \emph{affordance} (function or action possibility), \emph{state} (transient physical or operational condition), \emph{part-whole} (constituent or structural relation), \emph{negative} (explicit exclusion or absence), and \emph{compositional} (two or more jointly required semantic constraints). We use compositional queries when no single constituent constraint is sufficient to identify the target. Fig.~\ref{fig:envobench-pipeline}, panel~\Circled{3} provides examples of these $8$ prompt types.

\subsection{Target Annotation and Prompt Variants}\label{sec:target-annotation}
    We prepared Pro-Bench to support unconstrained natural language-conditioned robotic scene understanding. For each RGB frame (Section~\ref{sec:dataset}), we provide manual ground-truth annotations for both common and uncommon objects at the instance level using Roboflow annotation tools. Seven independent human annotators contributed fine-grained polygon boundaries, with an additional reviewer verifying correspondence between each annotation and its intended target. We used fine-grained polygon boundaries to minimise overlap between neighbouring instances and to support both box- and mask-based evaluation. We provide the ground-truth annotations in multiple formats, e.g., COCO~\cite{lin2014microsoft}, suitable for benchmarking visual reasoning models.

    To evaluate robustness to linguistic formulation, we associate each target with an original query $p^{\text{orig}}_{i}$. From $p^{\mathrm{orig}}_{i}$, we derive a shortened categorical form $p^{\mathrm{cat}}_{i}$ through an explicit, manually reviewed mapping to a concise class-level noun phrase that preserves the intended target while abstracting additional semantic constraints encoded in $p^{\mathrm{orig}}_{i}$ (e.g., attributes, relations, state, or affordance). 
    Furthermore, we construct the template-wrapped form as follows: $p^{\mathrm{tmpl}}_{i} =$ ``a photo of $\{p^{\mathrm{orig}}_{i}\}$''. This yields $P_{i} = \{p^{\mathrm{orig}}_{i}, p^{\mathrm{cat}}_{i}, p^{\mathrm{tmpl}}_{i}\}$, which we used to quantify the robustness to prompt specification, described in Section~\ref{sec:prompt-robustnes}.
    
\subsection{Prompt-Robustness and Spatial Stability}\label{sec:prompt-robustnes}
    \textbf{Prompt robustness.}
    We characterise how strongly the model's predictions depend on the prompts used to specify the target. Let $\mathcal{K}=\{\mathrm{cat},\mathrm{orig},\mathrm{tmpl}\}$, denote the three target-preserving prompt formulations (Section~\ref{sec:target-annotation}). We investigate four key questions: (RQ1) whether free-form semantic descriptions improve or degrade detection performance relative to short categorical labels; (RQ2) whether template wrapping (e.g., ``a photo of \{prompt\}'') affects grounding performance across different OVM architectures; (RQ3) whether a single prompting strategy generalises consistently across models; and (RQ4) whether target-preserving prompt reformulation alters target recovery and the spatial localisation of the recovered target.
    
    To quantify these RQ, let $m_k = \mathrm{mAP}^{(k)}$ denote the aggregate $\mathrm{mAP}_{.5}$ under formulation $k\in\mathcal{K}$. For RQ1, we directly measure the effect of replacing the categorical formulation with the original query as $\Delta_{\mathrm{orig-cat}} = \mathrm{m}_{\mathrm{orig}} - \mathrm{m}_{\mathrm{cat}}$, where positive values indicate that the additional semantic information contained in the original query improves detection performance, while negative values indicate otherwise. Similarly, for RQ2, we measure the effect of template wrapping as: $\Delta_{\mathrm{tmpl -orig}} = \mathrm{m}_{\mathrm{tmpl}} - \mathrm{m}_{\mathrm{orig}}$. Therefore, the maximum prompt-induced performance variation is defined as follows:
\begin{equation}
    \Delta_{\max} = \max_{k\in\mathcal{K}} \mathrm{m}_{k} - \min_{k\in\mathcal{K}}\mathrm{m}_{k},
\label{eq:worst_prompt_gap}
\end{equation}
    where larger $\Delta_{\max}$ indicates stronger sensitivity to prompt formulation.
    To address RQ3, we additionally record the best-performing formulation for each model.

\textbf{Target recovery and localisation consistency.}
    Aggregate performance metrics do not show whether different formulations recover the same physical target. We therefore introduce prompt-consistent recall (PCR) to quantify target-recovery consistency under prompt reformulations. For each target instance $g_{i}$, under the formulation $k$, we define the recovery indicator as shown:
\begin{equation}
    r_i^{(k)}(\tau)=
    \begin{cases}
    1, & \text{if }g_{i}\text{ is recovered at IoU}\geq\tau,\\
    0, & \text{otherwise}.
    \end{cases}
    \label{eq:recovery_indicator}
\end{equation}
    where $\tau$ is the IoU threshold, and $g_{i}$ is considered recovered if it is matched to a prediction for the corresponding query. Therefore, we compute the overall PCR as follows:
\begin{equation}
    \mathrm{PCR}_{\tau} = \frac{1}{N_c} \sum_{i=1}^{N_c} \prod_{k\in\mathcal{K}_i^{\mathrm{coref}}} r_i^{(k)}\bigr(\tau\bigl),
\label{eq:pcr}
\end{equation}
    where $N_c$ is the number of targets verified to preserve the target's intended referent, and $\mathcal{K}_i^{\mathrm{coref}}$ is their corresponding formulation set. In our experiments, we set $\tau = 0.5$. For target-level consistency analysis, only reformulations verified to preserve the intended referent are included in $\mathcal{K}_i^{\mathrm{coref}}$.
    
    Furthermore, to address RQ4, we evaluate whether target-preserving reformulations alter the spatial extent of the recovered prediction. For each target recovered under all $K_i=\mid\mathcal K_i^{\mathrm{coref}}\mid$ formulations, let $\hat B_i^{(k)}$ denote the prediction matched to the target $g_{i}$. We define instance-level prompt localisation consistency (PLC) as the mean pairwise IoU between prompt-conditioned predictions for targets recovered under all corresponding formulations as follows:
\begin{equation}
    \mathrm{PLC}
    =
    \frac{1}{|\mathcal{Q}^{*}|}
    \sum_{i\in\mathcal{Q}^{*}}
    \frac{2}{K_i(K_i-1)}
    \sum_{\substack{a<b}}
    \operatorname{IoU}
    \left(
    \hat{B}_{i}^{(a)},
    \hat{B}_{i}^{(b)}
    \right),
    \label{eq:plc_instance}
\end{equation}
where $\mathcal{Q}^{*}$ is the subset of targets recovered across all target-preserving formulations. Higher PLC indicates stronger spatial consistency in successful recovery.

\section{Experiments and Results}\label{sec:results}
    
\subsection{Models and Experimental Setup}\label{sec:setup}
    We evaluate $16$ representative open-vocabulary configurations with publicly hosted pre-trained weights\footnote{For ease of reproducibility and to avoid the API constraints, we only considered open-source OVMs with pre-trained weights available via Huggingface:~\url{https://huggingface.co/}, Ultralytics:~\url{https://docs.ultralytics.com/} or GitHub:~\url{https://github.com/}}, spanning transformer-based detectors, real-time single-stage models, and segmentation-based models (see Fig.~\ref{fig:pareto}). 
    We evaluated all the models in strict zero-shot, with frozen pre-trained weights and no Pro-Bench-specific fine-tuning.

    To preserve native inference behaviour, we follow each model's published prompt-conditioning interface. We also cached the models' predictions to ensure reproducible metric computation.
    All the experiments are conducted on the same NVIDIA RTX~4090 GPU (24~GB VRAM).

\subsection{Evaluation Protocol}\label{sec:metrics}
\textbf{Detection and localisation.}
    For each target query, we sort predictions by the model's confidence score from its inference and post-processing pipeline. We use these scores only to rank predictions within a given model-query pair. As in standard object-detection evaluation, we match predictions one-to-one with ground truth instances in confidence order at IoU threshold $\tau$. We report $\mathrm{mAP}_{.5}$, $\mathrm{mAP}_{.75}$, and $\mathrm{mAP}_{.5:.95}$, where $\mathrm{mAP}_{.5:.95}$ averages AP over $\tau\in\{0.5,0.55,\ldots,0.95\}$. To better characterise localisation, particularly for predictions that recover the correct semantic region but exhibit weak geometric alignment, we extend the IoU sweep to $\tau\in\{0.1,0.2,\ldots,0.9,0.95\}$.

    Although some models output masks, using mask AP as the main metric would restrict comparisons and prevent consistent benchmarking across architectures. We therefore use bounding-box AP as the primary localisation metric, since it applies uniformly to both detection- and segmentation-based OVMs. We nevertheless retain polygon annotations to support mask-based evaluation for segmentation-capable models and future Pro-Bench extensions.
    
\textbf{Inference latency.}
    We assess the models' computational efficiency via end-to-end inference latency, including preprocessing, the forward pass, and post-processing. For each model, we exclude three warm-up iterations and measure execution time over ten timed runs on a fixed subset of 50 images, using CUDA synchronisation before and after each inference on the same NVIDIA RTX 4090 GPU. Because prompt-conditioning and post-processing differ across architectures, we follow each model’s native inference protocol while holding the query workload constant, and report latency in the context of the model's specific prompt-processing protocol (Section~\ref{sec:setup}).

\textbf{Prompt robustness.}
    To evaluate the sensitivity to the prompt formulations, we directly employ the metrics described in Section~\ref{sec:prompt-robustnes}.
    
\subsection{Quantitative Results}\label{sec:results}
\begin{figure}[htp]
    \centering
    \includegraphics[width=1.0\linewidth]{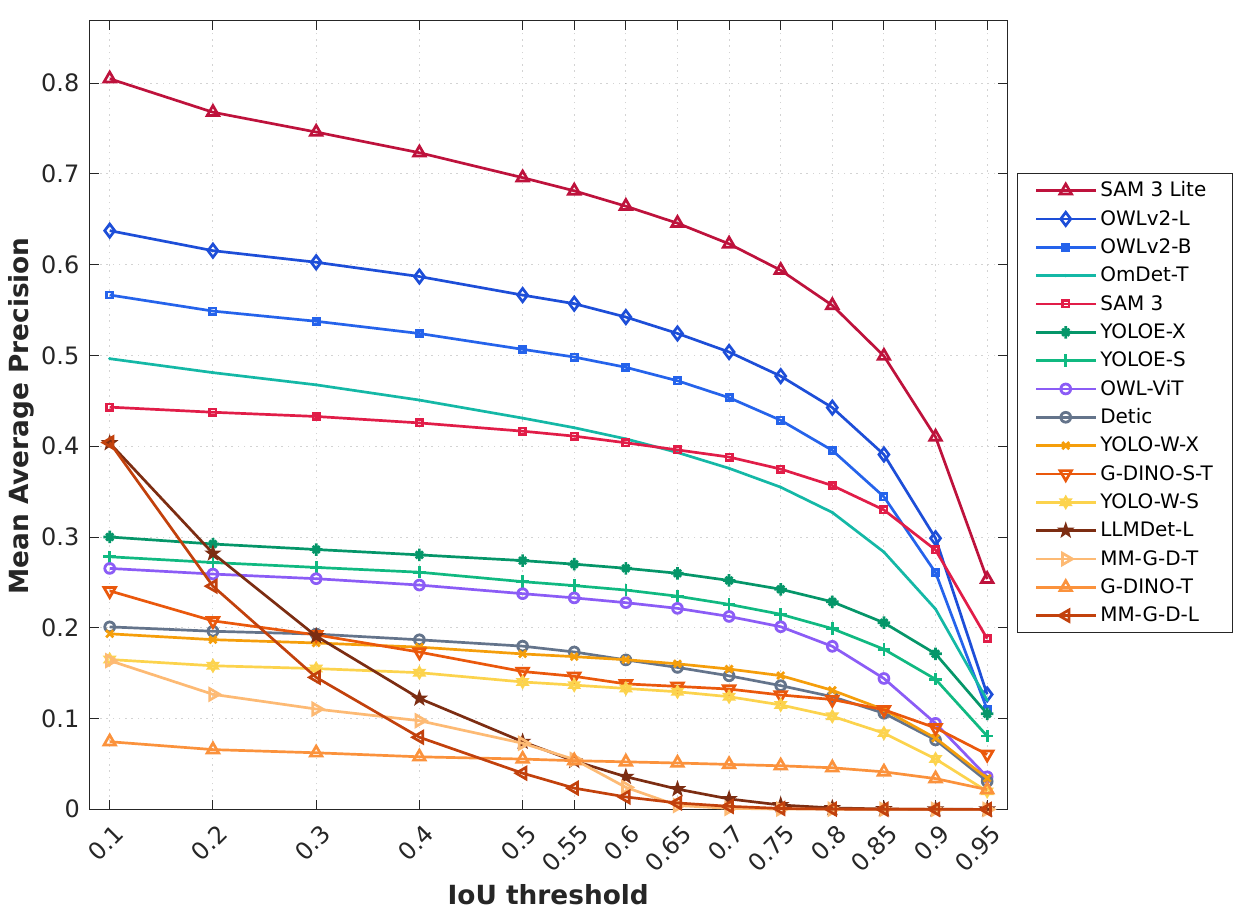}
    \caption{mAP versus IoU threshold across the benchmarked model configurations. The models' behaviours varied at stricter IoU thresholds, from moderate to sharp declines.}
    \label{fig:map_iou}
\vspace*{-0.50\baselineskip}
\end{figure}

\textbf{Detection and localisation performance.}
    Table~\ref{tab:quantitative-results} summarises the zero-shot detection, localisation, efficiency, and robustness to prompt variation across the OVM configurations, whereas Fig.~\ref{fig:map_iou} shows the performance across IoU thresholds. SAM3-Lite~\cite{zeng2026sam3} achieves the strongest overall localisation performance across the $\mathrm{mAP}_{.5}$, $\mathrm{mAP}_{.75}$, and $\mathrm{mAP}_{.5:.95}$. OWLv2-L~\cite{minderer2023scaling} and OWLv2-B~\cite{minderer2023scaling} follow at $\mathrm{mAP}_{.5}=0.567$ and $0.507$, respectively, while OmDet-T~\cite{zhao2024real} slightly surpasses SAM3~\cite{carion2025sam} at IoU $=0.5$. However, under stricter IoU thresholds, the behaviour changes: SAM3~\cite{carion2025sam} retains $85.1\%$ of its mAP$_{.5}$ under mAP$_{.5:.95}$, compared with $80.7\%$ for SAM3-Lite~\cite{zeng2026sam3} and $78.1\%$ for OWLv2-L~\cite{minderer2023scaling}. Fig.~\ref{fig:map_iou} further shows that several model configurations degrade sharply beyond IoU $= 0.5$. Thus, strong detection at IoU $= 0.5$ does not necessarily imply precise geometric localisation at stricter overlap thresholds.

\begin{table*}[htp]
\centering
\scriptsize
\caption{Quantitative Pro-Bench results. Top-five values are shaded from darkest (best) to lightest. $m_{\mathrm{orig}}$ corresponds to the reported $\mathrm{mAP}_{.5}$ row; $m_{\mathrm{cat}}$ and $m_{\mathrm{tmpl}}$ denote $\mathrm{mAP}_{.5}$ under the categorical and template-wrapped variants, respectively.}
\setlength{\tabcolsep}{2.0pt}
\renewcommand{\arraystretch}{1.05}
\begin{tabular}{@{}l c c c c c c c c c c c c c c c c@{}}
\toprule
 & 
\rotatebox{25}{\tiny \textbf{SAM3-Lite}} &
\rotatebox{25}{\tiny \textbf{OWLv2-L}} &
\rotatebox{25}{\tiny \textbf{OWLv2-B}} &
\rotatebox{25}{\tiny \textbf{SAM3}} &
\rotatebox{25}{\tiny \textbf{OmDet-T}} &
\rotatebox{25}{\tiny \textbf{YOLOE-X}} &
\rotatebox{25}{\tiny \textbf{YOLOE-S}} &
\rotatebox{25}{\tiny \textbf{OWL-ViT}} &
\rotatebox{25}{\tiny \textbf{G-DINO-S-T}} &
\rotatebox{25}{\tiny \textbf{YOLO-W-X}} &
\rotatebox{25}{\tiny \textbf{Detic}} &
\rotatebox{25}{\tiny \textbf{YOLO-W-S}} &
\rotatebox{25}{\tiny \textbf{G-DINO-T}} &
\rotatebox{25}{\tiny \textbf{LLMDet-L}} &
\rotatebox{25}{\tiny \textbf{MM-G-D-L}} &
\rotatebox{25}{\tiny \textbf{MM-G-D-T}} \\
\midrule

\textbf{$\mathrm{mAP}_{.5}$}
& \cellcolor{mAPgreenDark!100!white}\textbf{0.696}
& \cellcolor{mAPgreenDark!80!white}0.567
& \cellcolor{mAPgreenDark!60!white}0.507
& \cellcolor{mAPgreenDark!20!white}0.417
& \cellcolor{mAPgreenDark!40!white}0.431
& 0.274 & 0.251 & 0.238 & 0.152 & 0.171 & 0.180 & 0.140 & 0.055 & 0.075 & 0.040 & 0.073 \\

\textbf{$\mathrm{mAP}_{.75}$}
& \cellcolor{mAPgreenDark!100!white}\textbf{0.594}
& \cellcolor{mAPgreenDark!80!white}0.477
& \cellcolor{mAPgreenDark!60!white}0.429
& \cellcolor{mAPgreenDark!40!white}0.375
& \cellcolor{mAPgreenDark!20!white}0.355
& 0.242 & 0.215 & 0.201 & 0.126 & 0.147 & 0.136 & 0.115 & 0.048 & 0.005 & 0.001 & 0.000 \\

\textbf{$\mathrm{mAP}_{.5:.95}$}
& \cellcolor{mAPgreenDark!100!white}\textbf{0.562}
& \cellcolor{mAPgreenDark!80!white}0.443
& \cellcolor{mAPgreenDark!60!white}0.396
& \cellcolor{mAPgreenDark!40!white}0.355
& \cellcolor{mAPgreenDark!20!white}0.333
& 0.228 & 0.201 & 0.179 & 0.121 & 0.132 & 0.129 & 0.104 & 0.045 & 0.020 & 0.009 & 0.016 \\

\textbf{Lat. (ms)}
& 683.0 & 244.8 & 56.6 & 266.7
& \cellcolor{stableBlueDark!40!white}22.7
& 44.1
& \cellcolor{stableBlueDark!20!white}38.5
& \cellcolor{stableBlueDark!60!white}15.3
& 294.8
& \cellcolor{stableBlueDark!80!white}12.9
& 61.7
& \cellcolor{stableBlueDark!100!white}\textbf{7.3}
& 201.6 & 240.1 & 236.1 & 59.6 \\

\midrule
\textbf{$m_{\mathrm{cat}}$}
& \cellcolor{mAPgreenDark!100!white}\textbf{0.737}
& \cellcolor{mAPgreenDark!80!white}0.540
& \cellcolor{mAPgreenDark!60!white}0.492
& \cellcolor{mAPgreenDark!20!white}0.440
& \cellcolor{mAPgreenDark!40!white}0.454
& 0.188 & 0.182 & 0.257 & 0.126 & 0.216 & 0.190 & 0.167
& 0.037 & 0.108 & 0.041 & 0.078 \\

\textbf{$m_{\mathrm{tmpl}}$}
& \cellcolor{mAPgreenDark!80!white}0.541
& \cellcolor{mAPgreenDark!100!white}\textbf{0.569}
& \cellcolor{mAPgreenDark!60!white}0.509
& 0.044
& \cellcolor{mAPgreenDark!40!white}0.363
& 0.264
& \cellcolor{mAPgreenDark!20!white}0.299
& 0.234 & 0.177 & 0.163 & 0.174 & 0.134 & 0.124 & 0.030 & 0.039 & 0.002 \\

\textbf{$\Delta_{\mathrm{orig-cat}}$}
& $-0.041$ & $+0.027$ & $+0.015$ & $-0.023$
& $-0.022$ & $+0.086$ & $+0.069$ & $-0.019$
& $+0.026$ & $-0.045$ & $-0.010$ & $-0.027$
& $+0.019$ & $-0.033$ & $-0.002$ & $-0.005$ \\

\textbf{$\Delta_{\mathrm{tmpl-orig}}$}
& $-0.155$ & $+0.002$ & $+0.002$ & $-0.372$
& $-0.068$ & $-0.010$ & $+0.048$ & $-0.004$
& $+0.026$ & $-0.009$ & $-0.006$ & $-0.007$
& $+0.069$ & $-0.045$ & $-0.001$ & $-0.071$ \\

\textbf{$\Delta_{\max}$}
& 0.196
& \cellcolor{stableBlueDark!20!white}0.029
& \cellcolor{stableBlueDark!60!white}0.017
& 0.396
& 0.090
& 0.086
& 0.117
& \cellcolor{stableBlueDark!40!white}0.023
& 0.051
& 0.054
& \cellcolor{stableBlueDark!80!white}0.016
& 0.033
& 0.087
& 0.078
& \cellcolor{stableBlueDark!100!white}\textbf{0.002}
& 0.076 \\

\textbf{$k_m^\ast$}
& Cat. & Tmpl. & Tmpl. & Cat. & Cat. & Orig. & Tmpl. & Cat.
& Tmpl. & Cat. & Cat. & Cat. & Tmpl. & Cat. & Cat. & Cat. \\

\textbf{PCR$_{.5}$}
& \cellcolor{mAPgreenDark!100!white}\textbf{0.847}
& \cellcolor{mAPgreenDark!80!white}0.482
& \cellcolor{mAPgreenDark!60!white}0.426
& 0.024
& \cellcolor{mAPgreenDark!40!white}0.243
& 0.072
& 0.092
& \cellcolor{mAPgreenDark!20!white}0.180
& 0.082
& 0.089
& 0.147
& 0.075
& 0.026
& 0.007
& 0.006
& 0.001 \\

\textbf{PLC}
& 0.917
& 0.994
& 0.996
& 0.984
& 0.958
& 0.996
& 0.997
& \cellcolor{mAPgreenDark!100!white}\textbf{0.999}
& 0.965
& 0.967
& \cellcolor{mAPgreenDark!80!white}0.998
& 0.967
& 0.989
& 0.733
& 0.773
& 0.760 \\

\bottomrule
\end{tabular}
\par\smallskip
{\footnotesize
$k_m^\ast$ is the best-performing prompt formulation for each configuration.
\textbf{Model references:} SAM3-Lite~\cite{zeng2026sam3}, OWLv2-L/B~\cite{minderer2023scaling}, SAM3~\cite{carion2025sam}, OmDet-T~\cite{zhao2024real}, YOLOE-X/S~\cite{wang2025yoloe}, OWL-ViT~\cite{minderer2022simple}, G-DINO-S-T~\cite{liu2024grounding}, YOLO-W-X/S~\cite{cheng2024yolo}, Detic~\cite{zhou2022detecting}, G-DINO-T~\cite{liu2024grounding}, LLMDet-L~\cite{fu2025llmdet}, MM-G-D-L/T~\cite{zhao2024open}}.
\label{tab:quantitative-results}
\vspace*{-1.15\baselineskip}
\end{table*}

\textbf{Effect of prompt formulation (RQ1 - RQ3).}
    The evaluated prompt variants show strongly architecture-dependent sensitivity to linguistic reformulation. The original free-form queries improve $\mathrm{mAP}_{.5}$ over the categorical prompts for $6/16$ models but decrease it for $10/16$, with the largest gain for YOLOE-X~\cite{wang2025yoloe} ($+0.086$) and the largest drop for YOLO-W-X~\cite{cheng2024yolo} ($-0.045$). Template wrapping likewise has smaller but more asymmetric effects: $5/16$ model configurations improve relative to the original query, whereas $11/16$ degrade;  G-DINO-T~\cite{liu2024grounding} benefits most, while SAM3~\cite{carion2025sam} exhibits the largest degradation ($0.417 \rightarrow 0.044$). The SAM3~\cite{carion2025sam}'s large degradation demonstrates strong sensitivity to sentence-style or template-wrapped prompt formulation.

\begin{figure}
    \centering
    \includegraphics[width=1.0\linewidth]{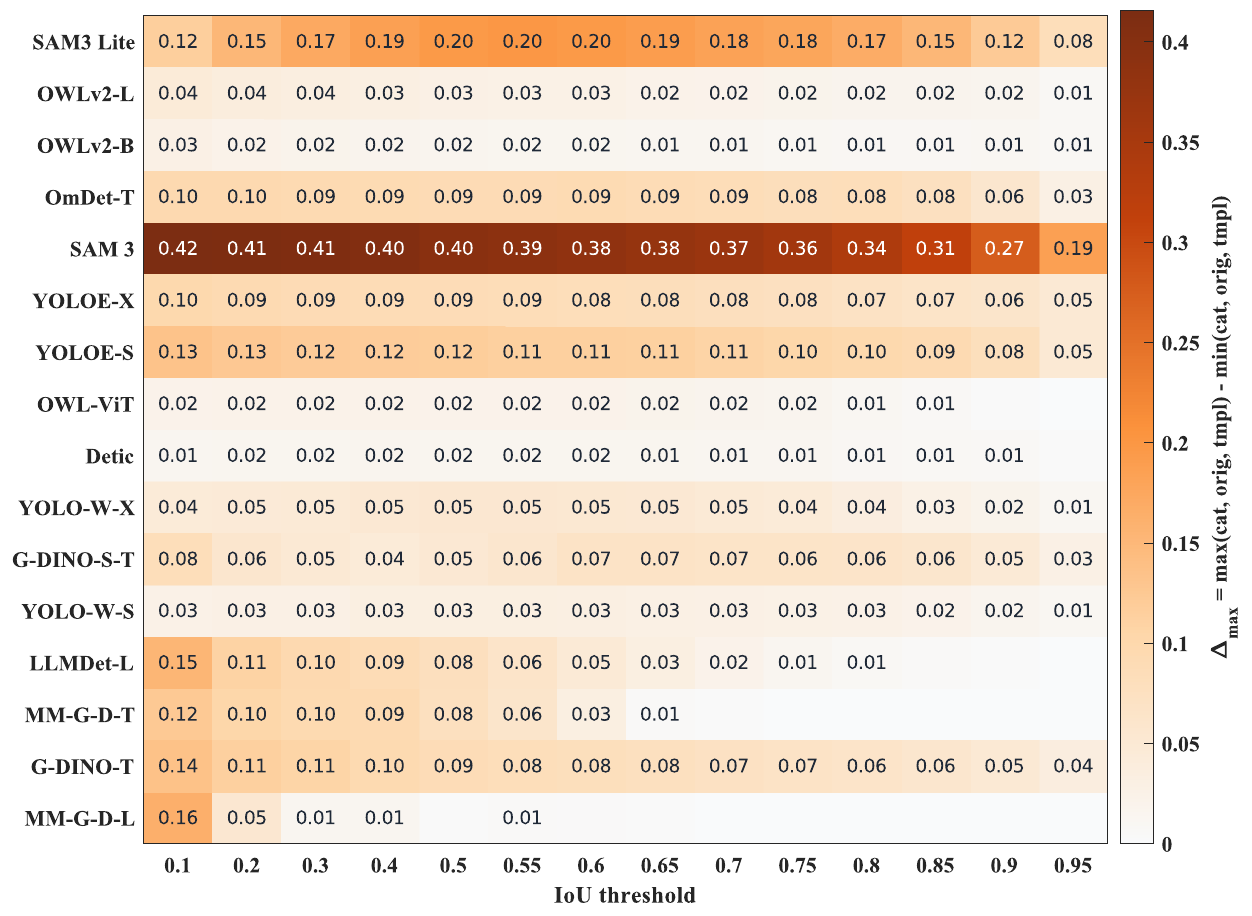}
    \caption{Prompt-induced performance variation versus IoU threshold across the model configurations. $\Delta_\mathrm{max}$ shrinks at strict IoU as the scores collapse toward zero. 
    }
    \label{fig:prompt-varaiation}
\vspace*{-1.15\baselineskip}
\end{figure}      
    Regarding RQ3, the best-performing formulation $k^{*}_{m}$ is the categorical form for $10/16$ configurations, the template-wrapped form for $5/16$, and the original free-form query for only YOLOE-X~\cite{wang2025yoloe}. Hence, although no single formulation is optimal across all the architectures, most current OVMs favour short categorical labels over the semantically richer queries that operators are likely to issue.
   $\Delta_{\max}$ (Table~\ref{tab:quantitative-results}) further quantifies this architecture dependent behaviour, and Fig.~\ref{fig:prompt-varaiation} shows its IoU-dependent performance variations.

\textbf{Target recovery and spatial consistency (RQ4).}
    PCR (Table~\ref{tab:quantitative-results}) reveals substantial differences in whether the same physical target remains recoverable under target-preserving prompt reformulations, which the aggregate mAP alone often conceals. SAM3-Lite~\cite{zeng2026sam3} exhibits the strongest recovery consistency, followed by the OWLv2 configurations~\cite{minderer2023scaling}, whereas several models recover only a small fraction of targets under all three formulations. Further, PLC provides a view of the successfully recovered subset: several configurations achieve very high spatial consistency despite substantially lower PCR. For example, SAM3~\cite{carion2025sam} attains high PLC ($0.984$) but very low PCR ($0.024$), indicating that its localisation is geometrically stable when recovery succeeds, although recovery itself is highly prompt-dependent. These results show that high spatial consistency among successfully recovered targets does not imply consistent target recovery.
    
\begin{figure}[t]
  \centering
  \includegraphics[width=\columnwidth]{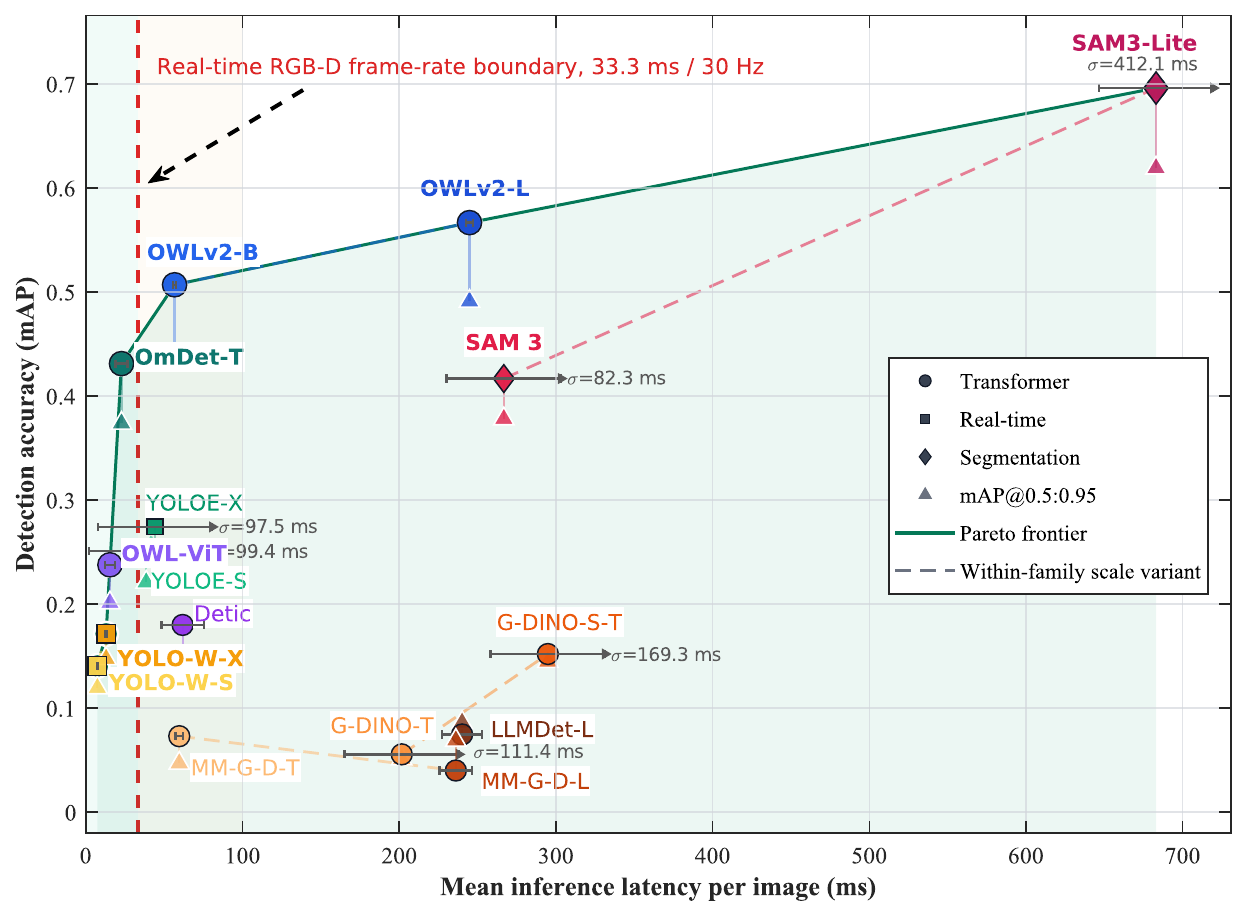}
  \caption{Accuracy-latency-localisation trade-off. We report the Pareto frontier to characterise the models' real-time performance. Vertical connectors: localisation tightness ($\mathrm{mAP_{0.5} \rightarrow mAP_{0.5:0.95}}$). Dashed connectors: scale variants of the same architecture. Horizontal lines: latency std. ($\pm \mathrm{ms}$). $0$-$33.3~\mathrm{ms}$: $30~\mathrm{Hz}$ real-time operational reference.}
  \label{fig:pareto}
  \vspace*{-1.15\baselineskip}
\end{figure}

\textbf{Accuracy-latency trade-off.}
    Fig.~\ref{fig:pareto} shows a clear trade-off between the grounding accuracy and the inference latency across all the model configurations.
    YOLO-W-S~\cite{cheng2024yolo} surprisingly is the fastest model achieving $7.3$\,ms per image, but unfortunately with a low grounding accuracy ($\mathrm{mAP}_{.5}=0.14$). Using the \(33.3\,ms\) (30\, Hz) as an operational real-time reference (e.g., standard Intel RealSense RGB-D camera frame rate), OmDet-T~\cite{zhao2024real} provides the strongest accuracy among the models, while YOLO-W-S/X~\cite{cheng2024yolo} occupy the ultra-low-latency region at much lower accuracy.

    Beyond this real-time reference, OWLv2-B/L~\cite{minderer2023scaling} offer better accuracy at increased computational cost, whereas SAM3-Lite~\cite{zeng2026sam3} achieves the strongest overall detection performance but with the highest latency.

\subsection{Qualitative Results}\label{sec:qualitative}
\begin{figure*}
    \centering
    \includegraphics[width=0.98\linewidth]{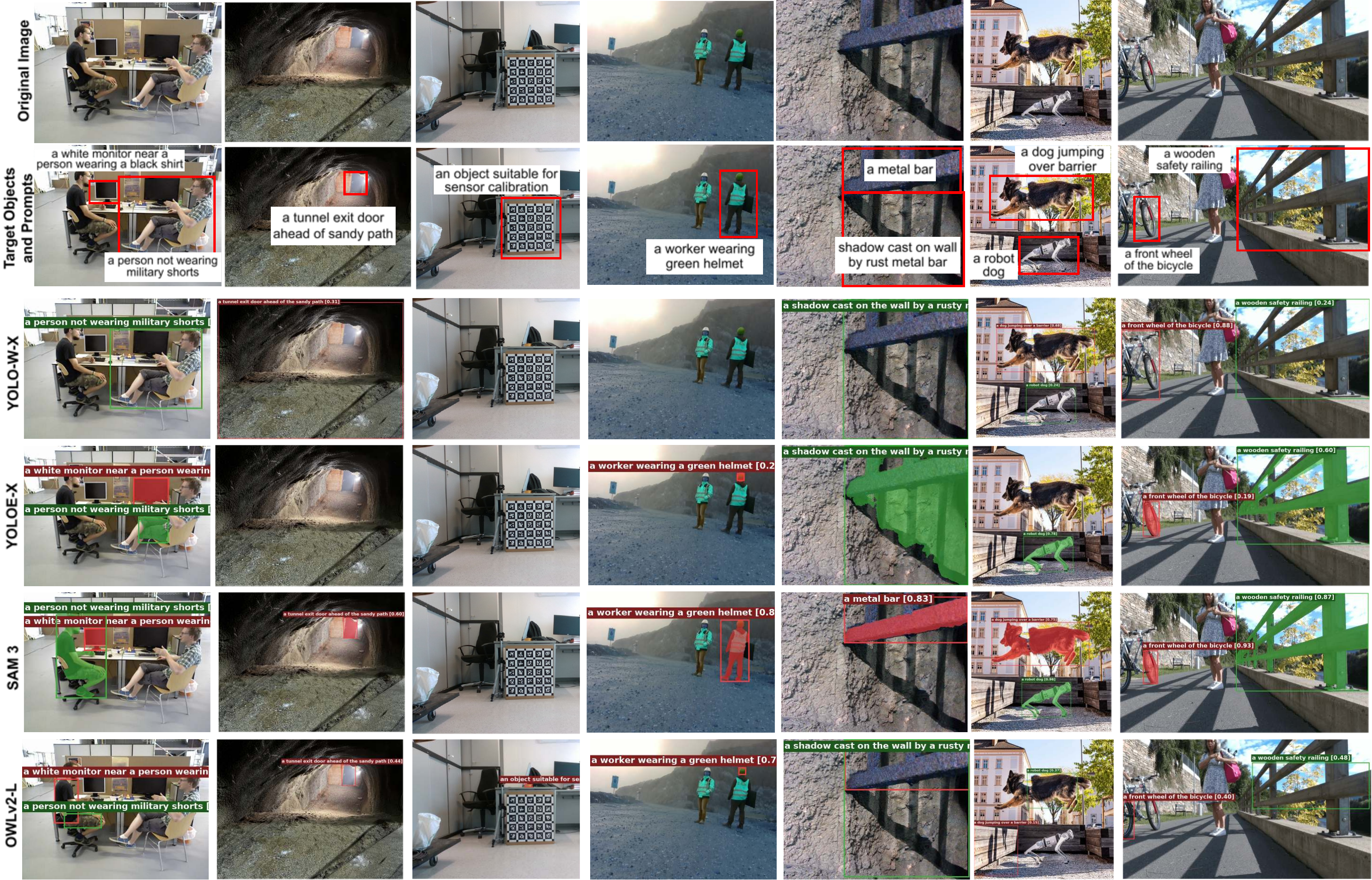}
    \caption{Qualitative grounding across representative Pro-Bench scenes. Predictions from the four representative OVM configurations representing real-time detection, open-vocabulary segmentation, and transformer-based behaviours. We performed the inference on the original images; outputs are resized only for visualisation, and the top-confidence prediction per target prompt is shown. More examples are provided on the project website:~\url{https://pro-bench.github.io/}.}
    \label{fig:qualitative}
    \vspace*{-1.15\baselineskip}
\end{figure*}

    Fig.~\ref{fig:qualitative} shows representative qualitative results across different Pro-Bench environments, prompt types, and target structures.
    It complements the quantitative evaluation by examining behaviours that are not fully characterised by the metrics in Table~\ref{tab:quantitative-results}, including partial localisation, spatial displacement, missed detections, and failures under semantically constrained queries. It further shows that similar aggregate mAP can conceal very different target-recovery and spatial localisation consistency under target-preserving reformulations. In particular, a model can recover the correct semantic region while localising it too coarsely. For the relational query ``a tunnel exit door ahead of the sandy path,'' OWLv2-L~\cite{minderer2023scaling} produces a compact prediction around the intended door, whereas YOLO-W-X~\cite{cheng2024yolo} responds to the same query with a substantially larger region spanning much of the tunnel. This shows how semantic recovery can coexist with weak geometric alignment, consistent with the IoU-dependent degradation shown in Fig.~\ref{fig:map_iou}.

    Further, visually salient targets can remain difficult when the query specifies the target indirectly. The affordance query ``an object suitable for sensor calibration'' yields no valid visible prediction from the four representative models (Fig.~\ref{fig:qualitative}) despite the calibration target being clearly visible in the scene. Overall, predictions diverge across architectures, showing that recognising a plausible object category does not guarantee that an OVM resolves the semantic constraints needed to identify the intended target.

\section{Conclusion}\label{sec:conclusion}
   In this work, we introduce Pro-Bench, a benchmark for open-vocabulary visual grounding under linguistically diverse, operator-style queries across heterogeneous real-world robotic environments. We evaluated $16$ representative open-vocabulary model configurations on the Pro-Bench dataset, and our results show that the majority of the model configurations perform best when conditioned on short categorical labels compared with free-form operator-style queries.  Moreover, the effect of prompt reformulation is strongly architecture-dependent, with template wrapping severely degrading some models' behaviour. PCR and PLC further show that consistent target recovery and localisation stability constitute distinct model properties: model configurations may exhibit highly stable localisation conditional on successful target recovery while nonetheless recovering only a limited fraction of targets across prompt reformulations. Overall, these results motivate the development of open-vocabulary models that are robust not only to visual-domain variation but also to how users linguistically specify the intended target. Our future work will extend Pro-Bench toward temporal video sequences, interaction-driven grounding, and a broader characterisation of prompt robustness in embodied settings.


\bibliographystyle{ieeetr}
\bibliography{references}


\addtolength{\textheight}{-12cm}   



\end{document}